\documentclass[runningheads]{llncs}
\usepackage[T1]{fontenc}
\usepackage{graphicx}
\usepackage{amsmath}
\usepackage{amssymb}
\usepackage{booktabs}
\usepackage{bm}
\usepackage{bbm}
\usepackage[misc]{ifsym}
\usepackage[pagebackref=true,breaklinks=true,colorlinks,bookmarks=false]{hyperref}
\begin{document}
\title{Intracranial Aneurysm Classification and Segmentation via Tri-Axial ROI and Multi-Task Learning}
\titlerunning{Intracranial Aneurysm Classification and Segmentation}
\author{Pengcheng Shi$^*$\inst{1} \and Kaiyuan Yang$^*$\inst{2} \and Houjing Huang$^*$\inst{2} \and \\ Jiawei Chen\inst{1} \and Yan Lu\inst{1} \and Jiaqi Liu\inst{1} \and Murong Xu\inst{2} \and \\ Minghui Zhang\inst{3} \and Bjoern Menze$^{(\textrm{\Letter})}$\inst{2} \and Xinglin Zhang$^{(\textrm{\Letter})}$\inst{1}}
\authorrunning{P. Shi et al.}
\institute{
    Medical Image Insights Co. Ltd., Shanghai, China
    \and University of Zurich, Zurich, Switzerland
    \and Shanghai Artificial Intelligence Laboratory, Shanghai, China
}
\maketitle
\footnotetext[1]{$^*$Equal contribution.}
\footnotetext[2]{$^{(\textrm{\Letter})}$Co-corresponding authors.}
\begin{abstract}
Intracranial aneurysms are often asymptomatic until rupture, which carries high mortality. Rupture risk assessment and treatment planning depend on both aneurysm morphology and anatomical location, yet existing automated methods remain limited to binary detection without fine-grained anatomical classification or multi-class segmentation. We present a multi-task framework that simultaneously performs multi-label classification, multi-class aneurysm segmentation, and multi-class vessel segmentation across 13 anatomical locations and four imaging modalities (CTA, MRA, T2, T1-post). Our two-stage approach combines a fast 2D tri-axial Region of Interest (ROI) extraction method with a 3D multi-task nnU-Net backbone. A dual-decoder design mitigates the extreme volume imbalance between aneurysm and vessel classes, while cross-attention pooling and modality-specific auxiliary heads improve feature learning across heterogeneous inputs. Our two-fold ensemble achieved 2nd place in the RSNA 2025 Intracranial Aneurysm Detection challenge. Code, model weights, a 3D Slicer plugin, and the corrected segmentation labels are publicly available.

\end{abstract}

%==============================================================================
\section{Introduction}
%==============================================================================

Intracranial aneurysms are pathological dilations of intracranial arteries. Rupture risk assessment and treatment planning depend on both aneurysm morphology and anatomical location. However, existing automated methods produce only binary outputs---aneurysm presence, aneurysm mask, and vessel mask, each binary---without addressing fine-grained anatomical classification or multi-class segmentation~\cite{hsu2025survey,timmins2021comparing} (Table~\ref{tab:method}). Beyond the limitation to binary outputs, the heterogeneity of imaging modalities presents additional difficulties. The RSNA 2025 Intracranial Aneurysm Detection (RSNA-2025) challenge~\cite{rsna-intracranial-aneurysm-detection} spans four modalities (CTA, MRA, T2, T1-post) which exhibit different contrast mechanisms, voxel spacings, and fields of view. Existing Region of Interest (ROI) extraction methods rely on modality-specific preprocessing or costly 3D sliding-window inference, limiting applicability at scale and efficient inference.

In this paper, we address these gaps with a two-stage framework. Stage~1 introduces a fast 2D tri-axial ROI extraction: sampling and segmenting three slices per orthogonal axis precisely localizes target vascular regions across modalities without full 3D inference. Stage~2 employs a 3D multi-task network built on nnU-Net~\cite{isensee2021nnu,isensee2024nnu} with a dual-decoder design that mitigates the extreme volume imbalance between aneurysm and vessel voxels. One decoder separates aneurysm from 13 vascular anatomy classes; the other maps each vessel anatomy with its aneurysm location class, enabling 26-class segmentation (13 vessel anatomy classes and 13 aneurysm location classes) alongside 13-location multi-label classification. Cross-attention pooling and a modality classification head further improve feature learning across heterogeneous inputs.

Our contributions are: (1) a 2D tri-axial ROI extraction strategy that handles multi-modal neurovascular data without full 3D segmentation, reducing inference to 9 slice predictions per volume; (2) a 3D multi-task framework with a dual-decoder design that unifies multi-label aneurysm classification, multi-class aneurysm segmentation, and multi-class vessel segmentation---the first to unify these three tasks at multi-class granularity across four modalities; and (3) ablation studies on the RSNA-2025 challenge~\cite{rsna-intracranial-aneurysm-detection} (2nd place) revealing that aneurysm segmentation quality, test-time augmentation (TTA), and ensembling each contribute to performance. Code, model weights, and segmentation checkpoints are publicly released.

\begin{table}[t]
\centering
\caption{Comparison of aneurysm analysis methods. Prior work produces only binary outputs; our method is the first to achieve multi-class outputs across all three dimensions. $\times$: not addressed. Aneu.: aneurysm; Cls.: classification; Seg.: segmentation.}
\label{tab:method}
\footnotesize
\setlength{\tabcolsep}{3pt}
\begin{tabular}{lcccc}
\toprule
\textbf{Study} & \textbf{Modality} & \textbf{Aneu. Cls.} & \textbf{Aneu. Seg.} & \textbf{Vessel Seg.} \\
\midrule
Ueda et al. (2019)~\cite{ueda2019deep} & MRA & Binary & $\times$ & $\times$ \\
Shi et al. (2020)~\cite{shi2020clinically} & CTA & Binary & Binary & $\times$ \\
Bo et al. (2021)~\cite{bo2021toward} & CTA & Binary & Binary & $\times$ \\
Di Noto et al. (2023)~\cite{di2023towards} & MRA & Binary & Binary$^\ddagger$ & $\times$ \\
Wei et al. (2024)~\cite{wei2024knowledge} & CTA & Binary & Binary & Binary \\
Yao et al. (2024)~\cite{yao2024aaseg} & CTA & Binary & Binary & Binary \\
\midrule
\textbf{Ours} & \shortstack{CTA, MRA, \\ T2, T1-post} & \textbf{13-location} & \textbf{13-location}$^\dagger$ & \textbf{13-anatomy}$^\dagger$ \\
\bottomrule
\multicolumn{5}{l}{\footnotesize $\dagger$Pseudo-label-based (active learning, Sec.~\ref{sec:data}). $\ddagger$Weak spherical labels.} \\
\end{tabular}
\end{table}

%==============================================================================
\section{Method}
\label{sec:method}
%==============================================================================

\begin{figure}[t]
\centering
\includegraphics[width=0.85\textwidth]{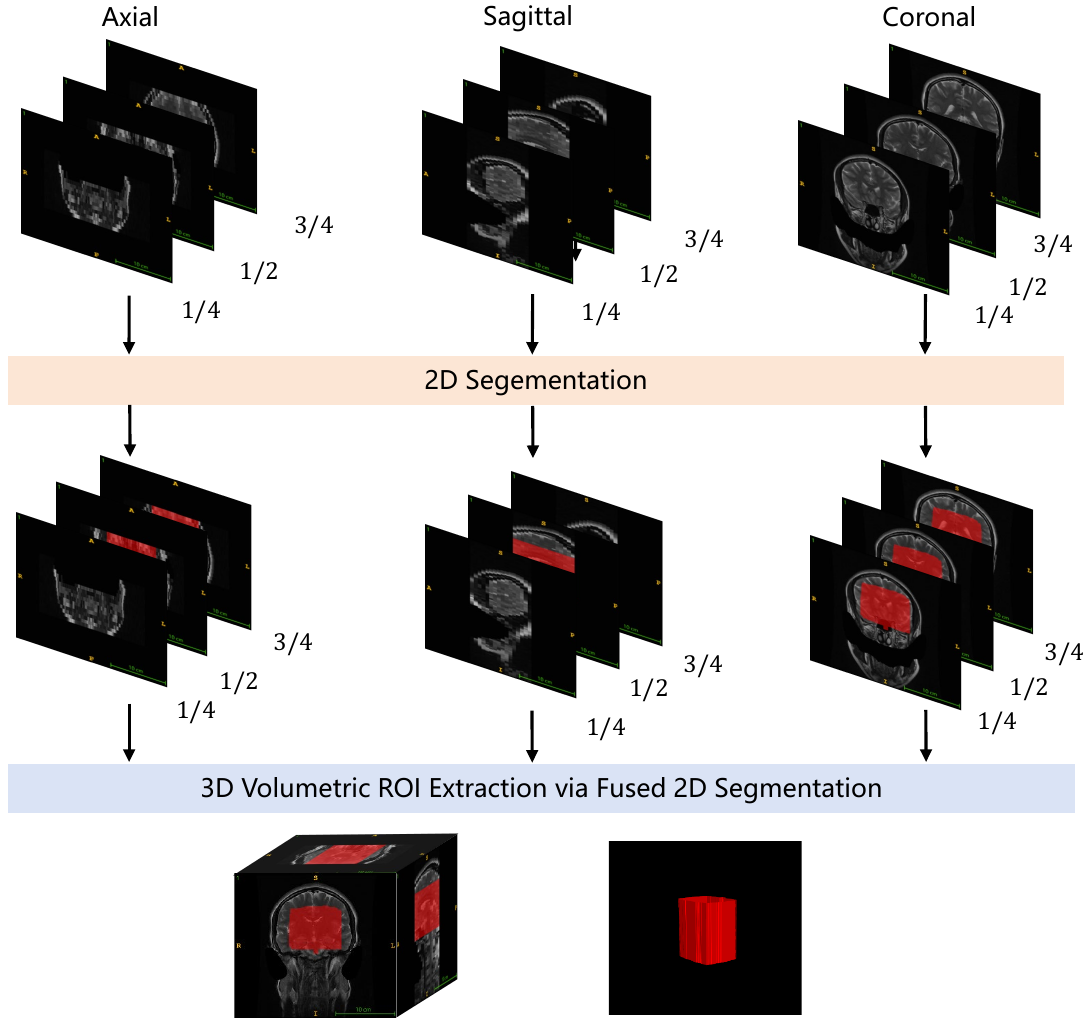}
\caption{Overview of the Stage~1 tri-axial ROI extraction pipeline. A 2D nnU-Net processes slices sampled from three orthogonal axes; bounding box coordinates are extracted from each 2D prediction and averaged across axes to obtain the final 3D ROI.}
\label{fig:framework}
\end{figure}

\begin{figure}[t]
    \centering
    \includegraphics[width=0.82\textwidth]{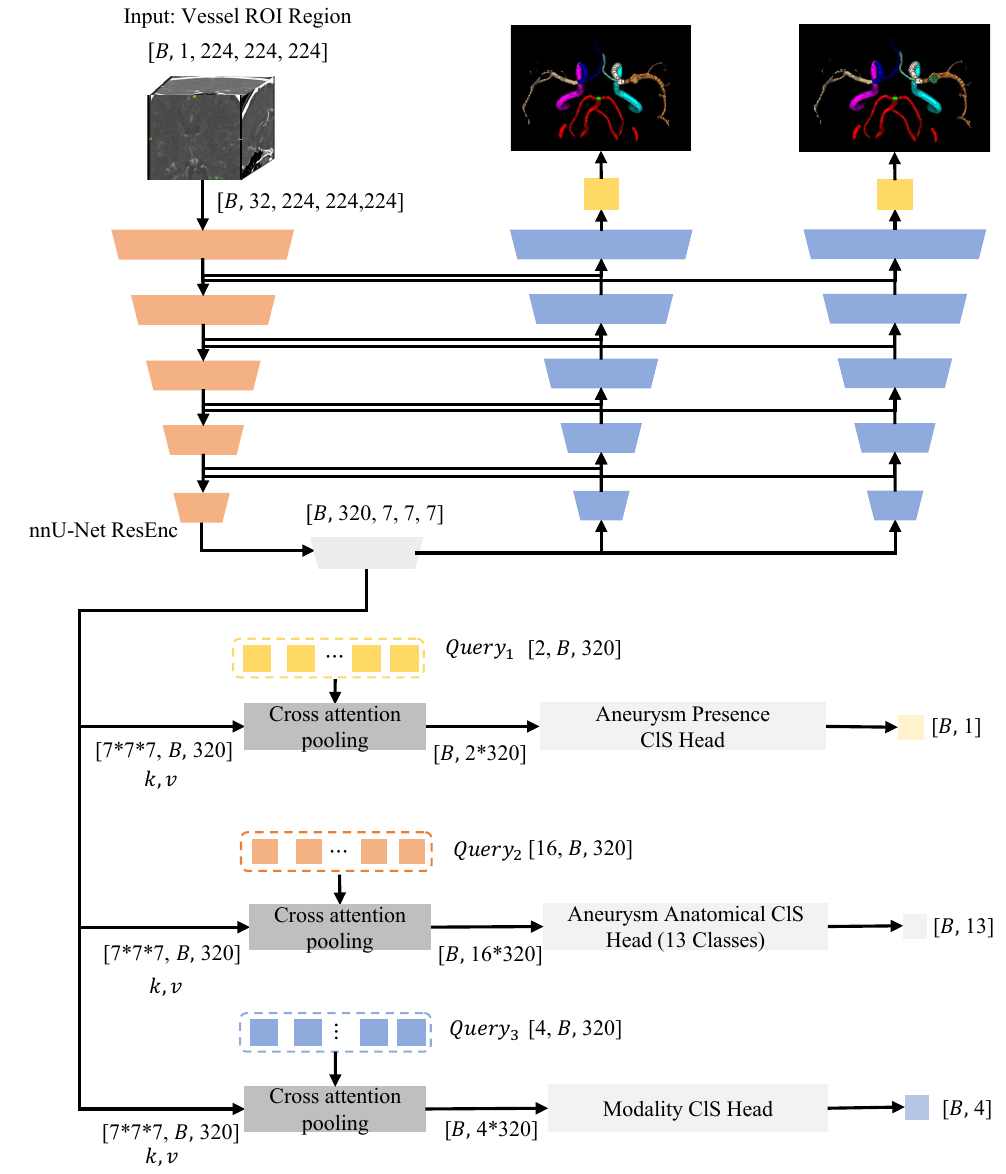}
    \caption{Stage~2 multi-task framework. Dual segmentation decoders produce vessel and aneurysm location masks; three classification heads predict modality, aneurysm presence, and location.}
    \label{fig:stage2}
\end{figure}

Fig.~\ref{fig:framework} illustrates Stage~1, which extracts a vascular ROI via 2D tri-axial inference. Stage~2 then performs joint segmentation and classification on the cropped volume, as shown in Fig.~\ref{fig:stage2}.

\subsection{Stage 1: 2D Tri-Axial ROI Extraction}
\label{sec:stage1}

To avoid the computational cost of 3D sliding-window inference, we propose a 2D tri-axial approach. For an input volume $\mathbf{X} \in \mathbb{R}^{C \times Z \times Y \times X}$, we extract 3 slices per axis at $\{1/4, 1/2, 3/4\}$ positions (up to 9 slices total), non-empty slices are resampled to in-plane spacing $[0.55, 0.5]$\,mm and processed by an nnU-Net 2D binary segmentation model~\cite{isensee2021nnu}.

Each 2D mask constrains two spatial dimensions---axial slices constrain $(y, x)$, sagittal constrain $(z, y)$, and coronal constrain $(z, x)$. Pooling predictions across axes yields the 3D bounding box via coordinate averaging. The final coordinates are:
\begin{equation}
    z_{\min} = \frac{1}{|S_Y \cup S_X|}\sum_{i \in S_Y \cup S_X} z_{\min}^i, \quad z_{\max} = \frac{1}{|S_Y \cup S_X|}\sum_{i \in S_Y \cup S_X} z_{\max}^i
\end{equation}
with $y$ analogously averaged over axial and sagittal predictions and $x$ over axial and coronal predictions, where $S_Y$ and $S_X$ denote the prediction sets from sagittal and coronal slices. The model is a standard nnU-Net 2D configuration trained on 1,090 cases spanning the RSNA-2025 challenge, TopCoW~\cite{yang2026topcow}, and TopBrain~\cite{yang2026topbrain}, covering diverse modalities and volume sizes. Training labels were generated by expanding vascular anatomy regions to a uniform physical size.

In local validation, extracted ROIs consistently encompassed the Circle of Willis and proximal arterial segments, covering nearly all 13 anatomical locations defined by the challenge dataset across all four modalities.

\subsection{Stage 2: 3D Multi-Task Learning}
\label{sec:stage2}

After Stage~1 cropping, volumes are resized to a uniform $224\times224\times224$ via trilinear interpolation. The network consists of a shared 6-stage Residual Encoder followed by two segmentation decoders and three classification heads (Fig.~\ref{fig:stage2}). Decoder~1 (14 output channels) separates 13 vascular anatomy classes from one merged aneurysm class; Decoder~2 (13 channels) pairs each vessel anatomy with its aneurysm location. The classification heads use cross-attention pooling on the encoder bottleneck: one predicts the input modality (4 classes, $Q{=}4$) as an auxiliary task, and two perform aneurysm classification---presence (1 output, $Q{=}2$) and multi-label location (13 outputs, $Q{=}16$).

The two decoders share the first 3 stages and diverge at the last 2. The original label space of 26 classes (13 vessel anatomies + 13 aneurysm locations) is remapped via index sets $\mathcal{I}_1$ and $\mathcal{I}_2$. $\mathcal{I}_1$ maps the 13 vessel anatomy classes individually plus all aneurysm location classes merged into one ($\{1\},\ldots,\{13\},\{14,\ldots,26\}$, 14 output channels). $\mathcal{I}_2$ pairs each vessel anatomy class with its corresponding aneurysm location following left-right anatomical symmetry ($\{1,26\},\{2,25\},\ldots,\{13,20\}$, 13 output channels). Post-training, the full 26-class segmentation is recovered: for each voxel, if Decoder~1 predicts the merged aneurysm class (class~14) and Decoder~2 predicts vessel anatomy class $k$, the final label is the aneurysm location class $k+13$; otherwise the voxel retains its vessel anatomy label $\hat{y}^{(2)}$.

\textbf{Dual-Decoder Design.} A single 26-class decoder suffers from extreme volume imbalance: aneurysm voxels are 2--3 orders of magnitude fewer than parent vessel voxels. In the 26-way softmax $\mathcal{L}_{\text{CE}} = -\frac{1}{N}\sum_{i,c} w_c \mathbbm{1}[y_i=c] \log(e^{z_{i,c}} / \sum_{j=1}^{26} e^{z_{i,j}})$, the denominator is dominated by large-volume vessel anatomy classes, suppressing aneurysm gradients regardless of per-class weights. Our dual-decoder design addresses this by (1)~halving the per-decoder softmax competitors (14 and 13 vs.\ 26), and (2)~consolidating all aneurysm signals into a single channel in Decoder~1 with a 10$\times$ CE loss weight, concentrating sparse gradients. Decoder~2's paired labels maintain anatomical correspondence with 5$\times$ CE weights on aneurysm-associated classes (compared to 1$\times$ for background and vessel classes).

\textbf{Multi-Task Loss.} The total loss is a weighted sum over all five heads:
\begin{equation}
    \mathcal{L} = \frac{1}{Z}\Big( \lambda_{\text{seg}} \sum_{d=1}^{2}\mathcal{L}_{\text{seg}d} \;+\; \mathcal{L}_{\text{BCE}}^{\text{presence}} \;+\; \mathcal{L}_{\text{BCE}}^{\text{location}} \;+\; \mathcal{L}_{\text{CE}}^{\text{modality}} \Big)
\end{equation}
Each $\mathcal{L}_{\text{seg}d}$ combines Dice and weighted CE loss~\cite{isensee2021nnu}, with $\lambda_{\text{seg}}=0.5$, $Z=3.5$. Aneurysm-associated classes receive higher CE weights (10$\times$ in Decoder~1, 5$\times$ in Decoder~2) to counter class imbalance; background and vessel anatomy classes use unit weight. BCE losses use inverse class frequency weighting.

\textbf{Training and Inference.} Models are trained with SGD and polynomial learning rate decay. The final submission ensembles two folds, each fine-tuned with refined segmentation masks (Sec.~\ref{sec:data}). Detailed training hyperparameters are provided in Sec.~\ref{sec:experiments}.

\begin{figure}[t]
    \centering
    \includegraphics[width=0.90\textwidth]{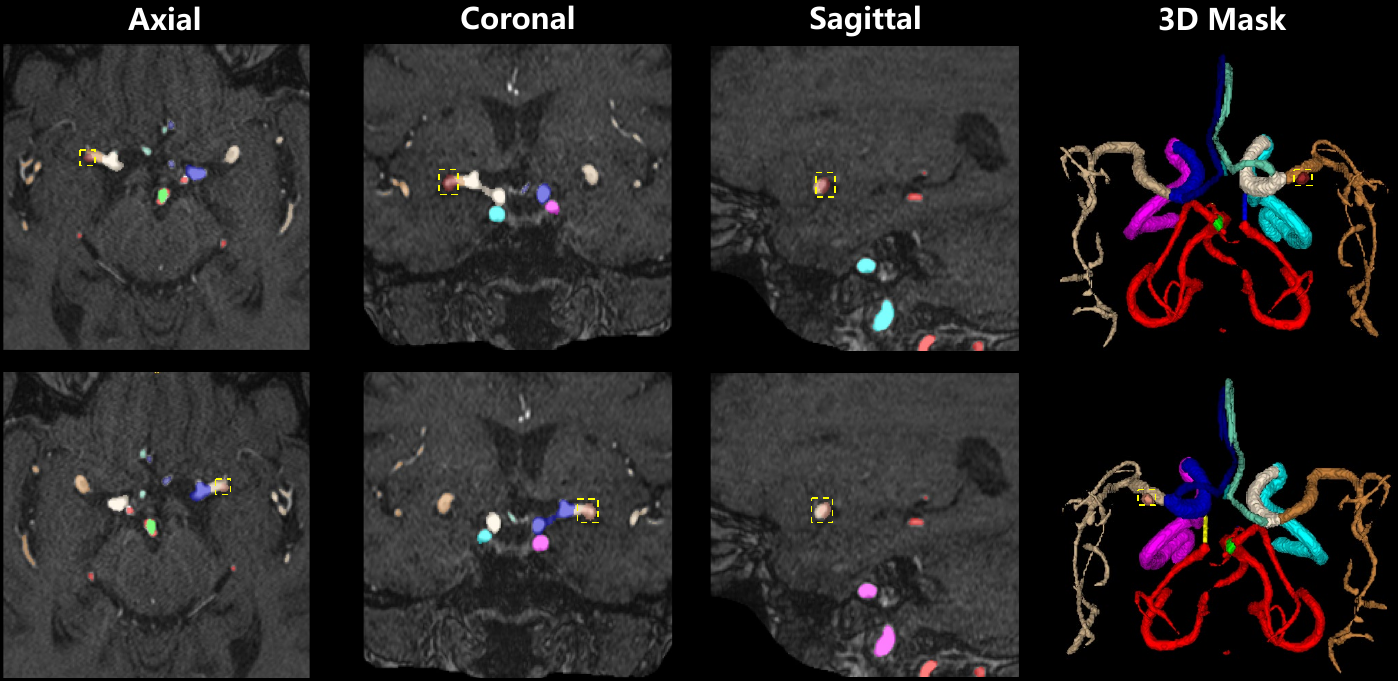}
    \caption{Left-right label-swap augmentation. Same case shown in original (top) and after horizontal flipping with swapped left-right labels (bottom), each in four views (axial, coronal, sagittal, 3D mask). Aneurysm regions are highlighted with yellow dashed bounding boxes.}
    \label{fig:vessel_flip}
\end{figure}

At inference, the cropped volume is resized to $224\times224\times224$, matching the patch size, and processed in a single forward pass. Decoders and the modality head are skipped for classification-only inference and included when segmentation output is required. We apply eight TTA augmentations (all axis-flip combinations), average classification logits across TTA variants and folds, and apply sigmoid for probabilities. Left-right-flipped location logits are corrected via the inverse label mapping (e.g., left middle cerebral artery to right middle cerebral artery). During training, the same label-swap is applied to paired left/right structures (Fig.~\ref{fig:vessel_flip}), doubling the dataset size for cases with left/right location labels.

%==============================================================================
\section{Experiments}
\label{sec:experiments}
%==============================================================================

\subsection{Data Preparation}
\label{sec:data}

The RSNA-2025 competition defines 13 aneurysm locations: left and right infraclinoid internal carotid artery (ICA), left and right supraclinoid ICA, left and right middle cerebral artery, left and right anterior cerebral artery, left and right posterior communicating artery, anterior communicating artery, basilar tip, and other posterior circulation. It provides 13-class vessel segmentation masks for 178 cases and aneurysm center-point coordinates for all annotated series, but not voxel-level aneurysm masks. We paired each of the 13 vessel classes with its corresponding aneurysm location class to form 26-class segmentation labels (13 vessel anatomies + 13 aneurysm locations), used as auxiliary supervision for the multi-task framework.

\textbf{Vessel anatomy masks.} Vessel anatomy supervision was built from three sources: the competition-provided Circle of Willis masks (178 cases), and voxel-level anatomical labels from TopCoW~\cite{yang2026topcow} (with LargeIA subset excluded) and TopBrain~\cite{yang2026topbrain}. To extend coverage to all 4,348 training series, we employed an active-learning approach inspired by TotalSegmentator~\cite{wasserthal2023totalsegmentator}: a model trained on the initial labels generated pseudo-labels for remaining cases, which were manually reviewed and corrected, then fed back into training iteratively.

\begin{figure}[t]
\centering
\includegraphics[width=0.95\textwidth]{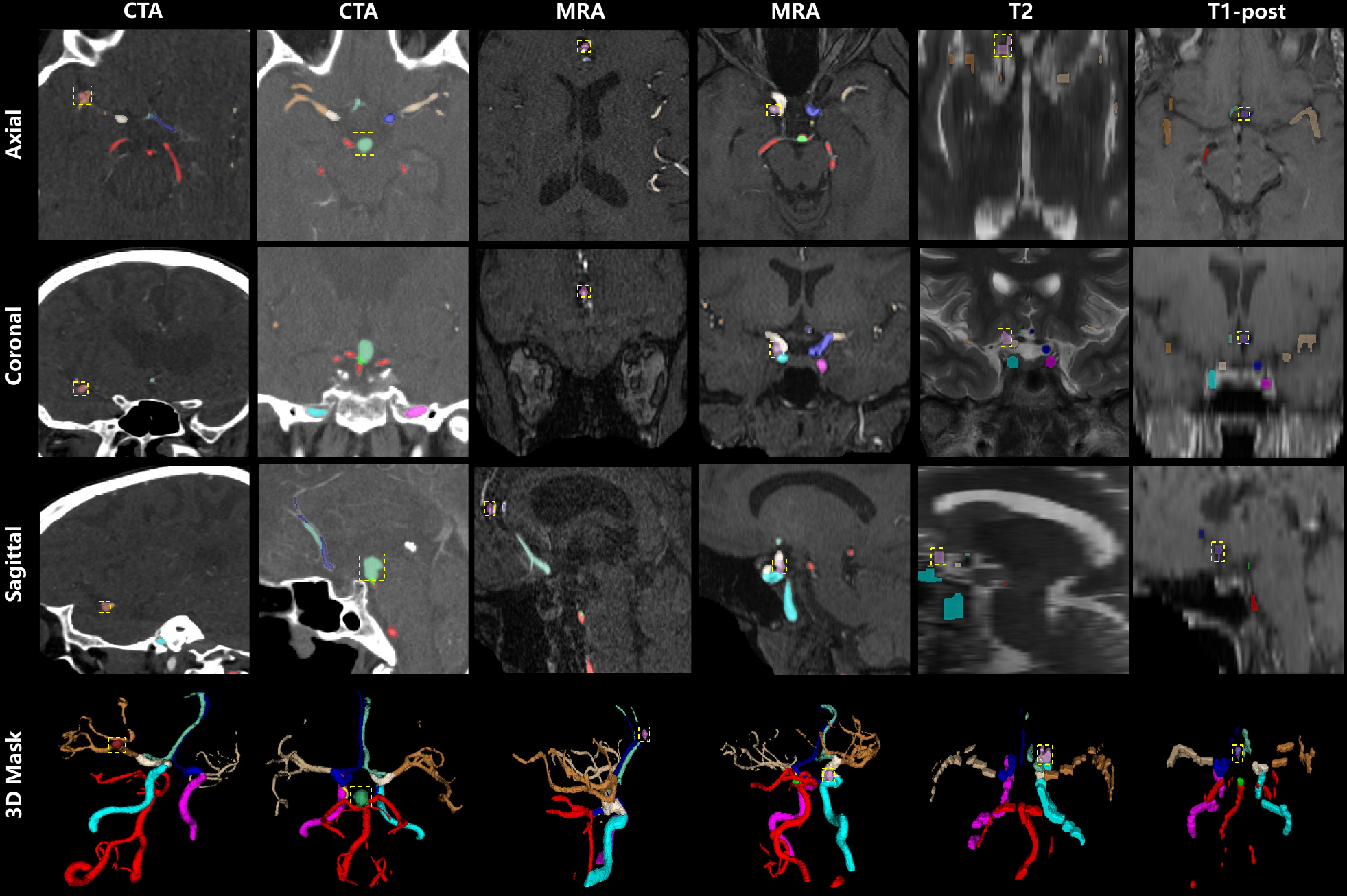}
\caption{Examples of the generated 26-class segmentation labels across modalities and anatomical locations. Each column depicts the same aneurysm in four views: axial slice with overlaid segmentation mask, coronal view, sagittal view, and 3D mask. Aneurysm regions are highlighted with yellow dashed bounding boxes.}
\label{fig:data_examples}
\end{figure}

\textbf{Aneurysm mask generation.} Aneurysm masks were generated from the competition-provided center-point coordinates through the same active-learning approach. Initial masks were created by placing 3D bounding boxes (2--5\,mm per side) around each center-point coordinate, then used to train an intermediate 14-class segmentation model (13 vessel anatomies + 1 aneurysm). The model was applied to all training cases; each predicted aneurysm region was assigned to an anatomical location via center-point distance and connected-component analysis. Predictions were manually corrected and fed back into training. This predict--correct--retrain cycle was repeated until all 4,348 series were covered. Fig.~\ref{fig:data_examples} shows representative examples of the final labels across modalities and anatomical locations. The corrected 26-class vessel-anatomy and aneurysm segmentation labels are publicly released in the original image space at \url{https://huggingface.co/datasets/spc819/rsna2025-aneurysm-26class-seg}; these corrected masks were produced by our team during the competition period for the competition ranking, have not been clinically reviewed, and are shared with the research community for further study.

\textbf{Label correction.} Manual review identified annotation errors in the original RSNA labels, including left/right side confusion and supra-/infra-clinoid ICA misclassifications, which were corrected where identified.

These corrections were necessary for the multi-class segmentation task, since incorrect anatomical labels would directly conflict with vessel anatomy supervision and degrade the dual-decoder training signal.

\subsection{Dataset and Setup}

We use all 4,348 RSNA-2025 series~\cite{rsna-intracranial-aneurysm-detection} for 5-fold cross-validation with random series-level splits, and select the best two folds for the final ensemble submission. Stage~1 follows the standard nnU-Net 2D pipeline~\cite{isensee2021nnu} with a custom multi-axial inference wrapper; Stage~2 extends the nnU-Net 3D pipeline with multi-task heads and TTA-based classification inference. Training uses SGD with initial learning rate 0.01, weight decay ($3{\times}10^{-5}$), and batch size 2 on a single NVIDIA A100 (80\,GB). Each model is trained for 1,000 epochs and fine-tuned for 100 or 250 epochs at reduced learning rate ($\eta_0{=}4{\times}10^{-3}$). Evaluation follows the competition protocol: public leaderboard (32\% of test data) and private leaderboard (68\%). The official metric computes the area under the ROC curve (AUC) for each of the 14 output labels---aneurysm presence plus 13 anatomical locations---then averages them with a weight of 13 on the presence label and 1 on each location label, equivalent to averaging the Aneurysm Presence AUC ($\mathrm{AUC}_{\text{AP}}$) and the mean of the 13 location AUCs: $\frac{1}{2}\big(\mathrm{AUC}_{\text{AP}} + \frac{1}{13}\sum_{i=1}^{13}\mathrm{AUC}_i\big)$.

\subsection{Ablation Study}

The ablation study (Table~\ref{tab:ablation}) tracks the progression of our solution. The Baseline uses initial V1 aneurysm masks from center-point coordinates. The ``Improved Data'' row replaces these with refined masks alongside external vessel anatomy labels from TopCoW and TopBrain; due to the cumulative design, this row reflects the joint effect of annotation refinement and external data expansion. Refined annotations yield gains of +0.018 public and +0.032 private; TTA and ensembling provide further improvements (4$\times$: +0.026/+0.012 public/private; 8$\times$: +0.010/+0.005; ensemble: +0.002/+0.005). The persistent 3--4\% public-private gap suggests uneven distribution of challenging cases across test splits.

\begin{table}
\centering
\caption{Ablation study on the RSNA 2025 challenge. TTA: Test-Time Augmentation. Public/Private: AUC on the public/private leaderboard.}
\label{tab:ablation}
\footnotesize
\setlength{\tabcolsep}{3pt}
\begin{tabular}{l|c|c|l}
\toprule
\textbf{Configuration} & \textbf{Public} & \textbf{Private} & \textbf{Notes} \\
\midrule
Baseline & 0.84407 & 0.81268 & \scriptsize V1 annotation, Fold~0, cls\_modality \\
+ TTA 4$\times$ & 0.87056 & 0.82508 & \scriptsize 4$\times$ TTA \\
+ Improved Data + TTA 4$\times$ & 0.88832 & 0.85718 & \scriptsize Refined seg. annotations \\
+ Improved Data + TTA 8$\times$ & 0.89805 & 0.86228 & \scriptsize 8$\times$ TTA, left-right label swap \\
+ Ensemble (2-fold) & \textbf{0.90035} & \textbf{0.86727} & \scriptsize Fold~0 (100 ep.) + Fold~1 (250 ep.) \\
\bottomrule
\end{tabular}
\end{table}

\subsection{Inference Speed}

We benchmarked the Stage~1 inference speed of our 2D tri-axial ROI against nnU-Net 3D sliding-window inference on 100 external cases using an NVIDIA RTX 3090. The 2D approach achieves a 12.7$\times$ average speedup (0.51\,s vs.\ 6.44\,s per volume), with the 3D model using 1\,mm isotropic spacing with patch size $224\times224\times224$. The 2D model processes individual slices at in-plane spacing $[0.55,0.5]$\,mm with patch size $320\times448$.

\subsection{3D Slicer Plugin}

\begin{figure}[ht]
\centering
\begin{minipage}{0.48\textwidth}
\centering
\includegraphics[width=\textwidth]{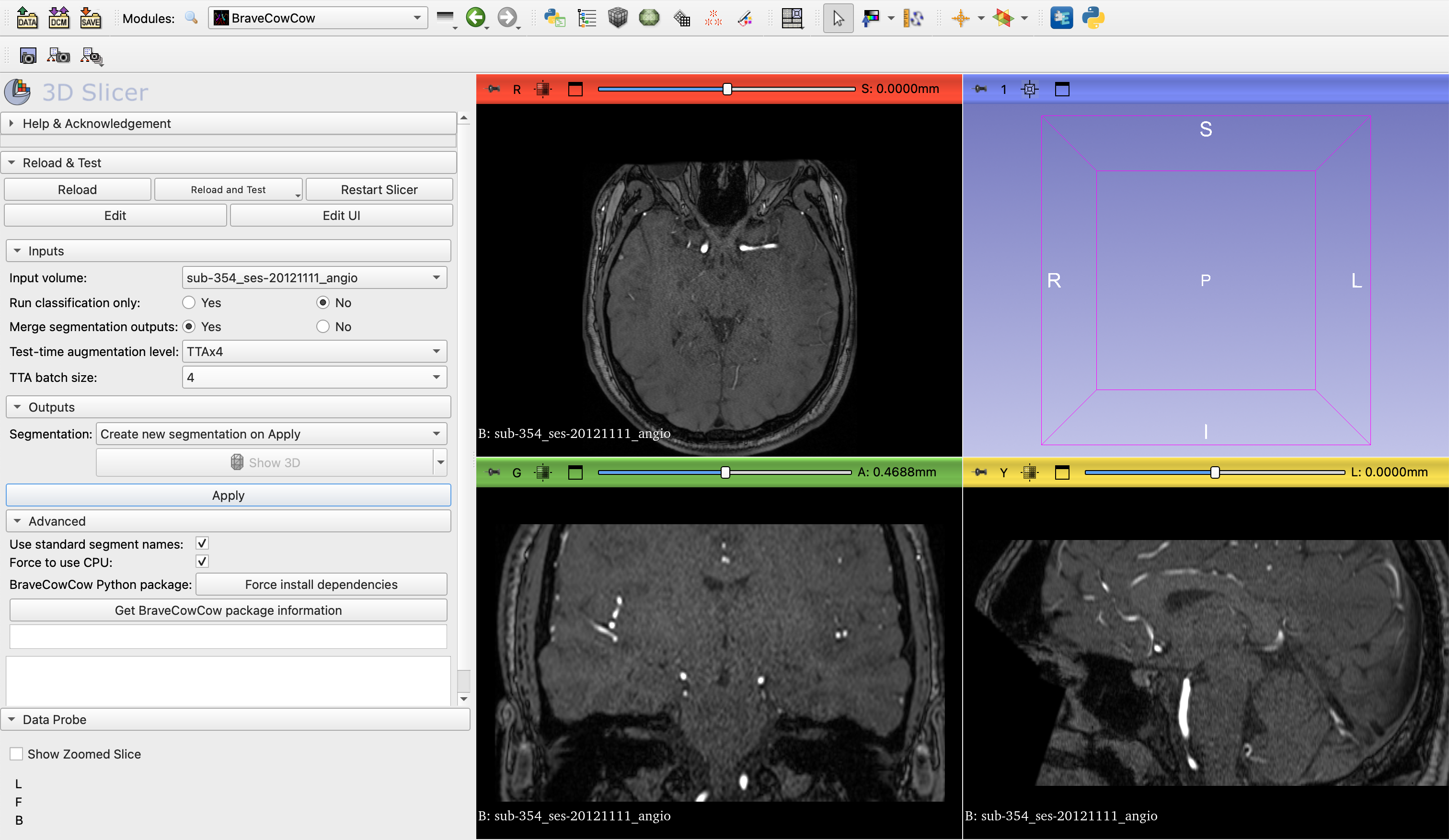}
\\[-2pt]
\textbf{(a)}
\end{minipage}
\hfill
\begin{minipage}{0.48\textwidth}
\centering
\includegraphics[width=\textwidth]{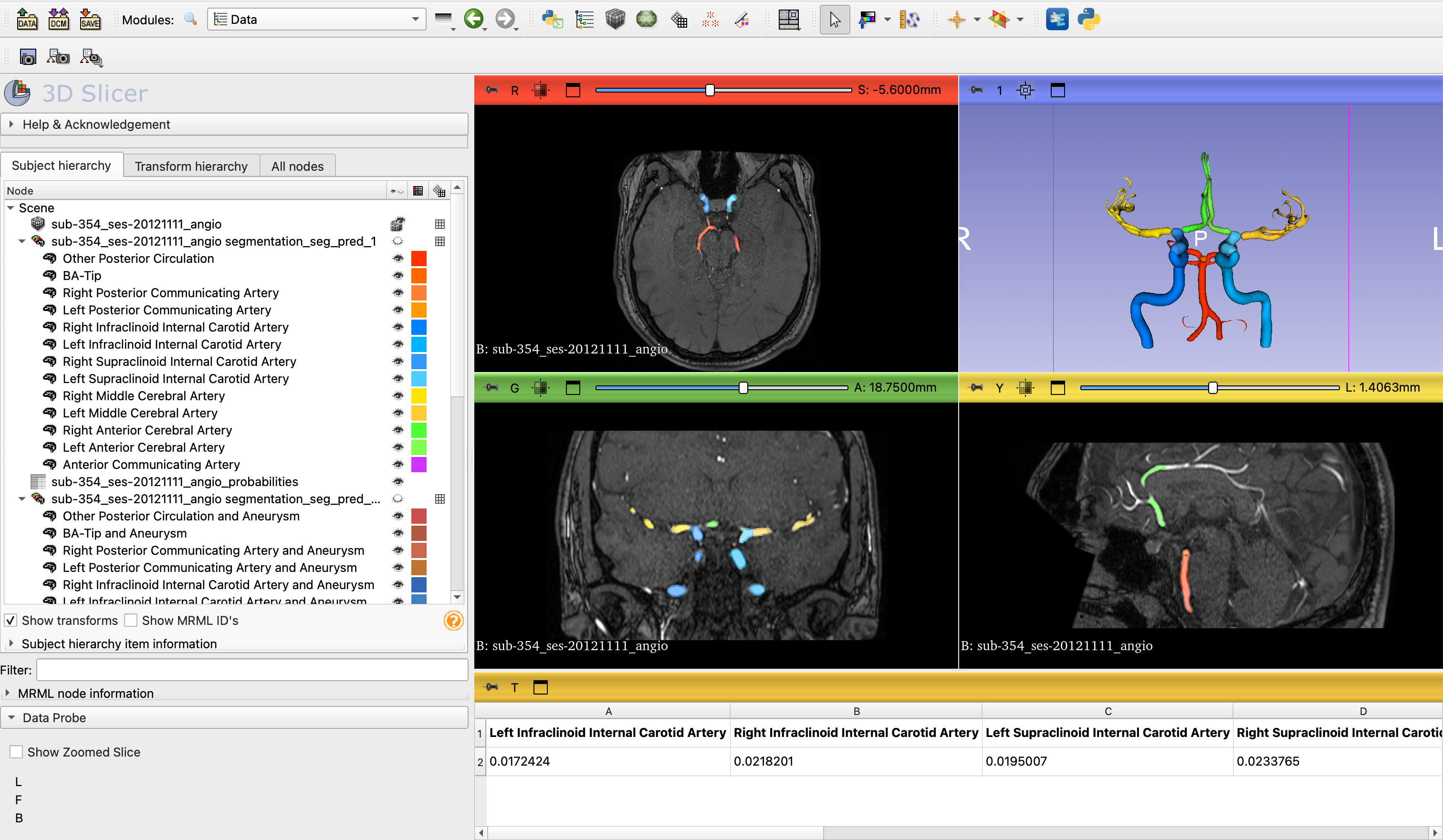}
\\[-2pt]
\textbf{(b)}
\end{minipage}
\caption{(a)~Plugin interface within 3D Slicer. (b)~Segmentation and classification results on an MRA scan.}
\label{fig:slicer}
\end{figure}

For clinical and research use, we packaged the two-stage framework as a 3D Slicer extension (Fig.~\ref{fig:slicer}), supporting all four modalities (CTA, MRA, T2, T1-post) with configurable TTA levels (1$\times$, 4$\times$, 8$\times$) and three output modes (14-class (vessel + merged aneurysm), 13-class merged, 26-class vessel+aneurysm). Source code is at \url{https://github.com/murong-xu/SlicerBraveCowCow} with the inference backend at \url{https://github.com/huanghoujing/bravecowcow_inference_docker}.

%==============================================================================
\section{Discussion}
%==============================================================================

\textbf{Limitations.} First, the challenge provides only center-point aneurysm annotations, so segmentation evaluation remains qualitative (Fig.~\ref{fig:data_examples}). Second, aneurysm and vessel anatomy pseudo-labels were generated via active learning~\cite{wasserthal2023totalsegmentator} and have not been systematically benchmarked against independent expert annotations. Third, the 13-class aneurysm segmentation is not fully end-to-end, requiring post-processing to fuse the outputs of two decoders, which introduces additional complexity and leaves room for architectural simplification.

Despite these limitations, multi-task learning with auxiliary segmentation heads produced dense anatomical outputs even when trained primarily for classification. The dual-decoder design offers a strategy for handling class imbalance where target aneurysm structures exhibit extreme volume disparities relative to surrounding vessel class. The fast 2D tri-axial ROI approach achieves efficient region extraction by exploiting the redundancy of multi-axis sampling, decoupling inference cost from volume size---even when the 3D baseline uses 1\,mm isotropic resolution.

%==============================================================================
\section{Conclusion}
%==============================================================================

We presented a two-stage framework that unifies multi-label aneurysm classification, multi-class aneurysm segmentation, and multi-class vessel segmentation across four modalities. The 2D tri-axial ROI extraction strategy enables fast region localization without full 3D inference, achieving a 12.7$\times$ average speedup over 3D sliding-window inference, while the dual-decoder design addresses the extreme volume imbalance between aneurysm and vessel classes. Ablation reveals that annotation quality, TTA, and ensembling each contribute to performance; future work should focus on improving segmentation annotation quality. We release code, model weights, and segmentation checkpoints to support further work.

\subsubsection{Code and Model Availability.}
\begin{sloppypar}
Code is available at \url{https://github.com/PengchengShi1220/RSNA2025_Intracranial-Aneurysm-Detection}. Stage~1 ROI model (\url{https://www.kaggle.com/models/pengchengshi/dataset180_2d_vessel_box_seg_stable}) and Stage~2 model weights (\url{https://www.kaggle.com/models/pengchengshi/dataset660_26classes_resize224_4661}) are available at Kaggle Models. The 3D Slicer extension is available at \url{https://github.com/murong-xu/SlicerBraveCowCow} with the inference backend at \url{https://github.com/huanghoujing/bravecowcow_inference_docker}.
\end{sloppypar}

\subsubsection{Data Availability.}
\begin{sloppypar}
The challenge dataset is available at \url{https://www.kaggle.com/competitions/rsna-intracranial-aneurysm-detection/data}. TopCoW (\url{https://doi.org/10.5281/zenodo.15692630}) and TopBrain (\url{https://doi.org/10.5281/zenodo.16878417}) annotations are available on Zenodo. The corrected 26-class vessel-anatomy and aneurysm segmentation labels used in this work are publicly released at \url{https://huggingface.co/datasets/spc819/rsna2025-aneurysm-26class-seg}. These corrected masks were produced by our team during the competition period for the competition ranking and have not been clinically reviewed; we share them with the research community for further study.
\end{sloppypar}

\subsubsection{Acknowledgements.}
The authors thank Medical Image Insights and UZH for compute support, and the Helmut Horten Foundation for funding support. We are grateful to the RSNA/Kaggle challenge organizers, the nnU-Net developers, and Kaggle forum contributors.

\subsubsection{Disclosure of Interests.}
The authors have no competing interests to declare that are relevant to the content of this article.

%==============================================================================
% Bibliography
%==============================================================================
\bibliographystyle{splncs04}
\bibliography{ref}

\end{document}